\documentclass[conference]{IEEEtran}
\IEEEoverridecommandlockouts

\usepackage[T1]{fontenc}

\usepackage{cite}
\usepackage{amsmath,amssymb,amsfonts}
\usepackage{graphicx}
\usepackage{textcomp}
\usepackage{xcolor}
\usepackage{hyperref}
\usepackage{subcaption}
 \usepackage[pscoord]{eso-pic}

\def\figurename{Fig.}

\makeatletter
\def\ps@IEEEtitlepagestyle{%
  \def\@oddhead{%
    \hfill
    \parbox[b]{.72\textwidth}{\raggedleft\footnotesize
      IEEE 2nd International Conference on Computing, Applications and Systems (COMPAS 2025)\\
      23--24 October 2025, Kushtia, Bangladesh%
    }%
  }%
  \def\@oddfoot{}%
}
\makeatother

\title{Unsupervised Anomaly Detection for Smart IoT Devices: Performance and Resource Comparison}

\author{\IEEEauthorblockN{Md. Sad Abdullah Sami}
\IEEEauthorblockA{\textit{Department of Electrical and Electronic Engineering} \\
\textit{Bangladesh University of Engineering and Technology}\\
Dhaka, Bangladesh \\ sadabdullahsami47@gmail.com}
\and
\IEEEauthorblockN{Mushfiquzzaman Abid}
\IEEEauthorblockA{\textit{Department of Electrical and Electronic Engineering} \\
\textit{Bangladesh University of Engineering and Technology}\\
Dhaka, Bangladesh \\ abid.mushfiq@gmail.com}
}

\begin{document}

\maketitle


\newcommand{\placetextbox}[3]{
 \setbox0=\hbox{#3}
 \AddToShipoutPictureFG*{ \put(\LenToUnit{#1\paperwidth},\LenToUnit{#2\paperheight}){\vtop{{\null}\makebox[0pt][c]{#3}}}
 }
 }
 \placetextbox{.23}{0.055}{\small{979-8-3315-5525-2/25/\$31.00~\copyright 2025 IEEE}}

\begin{abstract}
The rapid expansion of Internet of Things (IoT) deployments across diverse sectors has significantly enhanced operational efficiency, yet concurrently elevated cybersecurity vulnerabilities due to increased exposure to cyber threats. Given the limitations of traditional signature-based Anomaly Detection Systems (ADS) in identifying emerging and zero day threats, this study investigates the effectiveness of two unsupervised anomaly detection techniques, Isolation Forest (IF) and One-Class Support Vector Machine (OC-SVM), using the TON\_IoT thermostat dataset. A comprehensive evaluation was performed based on standard metrics (accuracy, precision, recall, and F1-score) alongside critical resource utilization metrics such as inference time, model size, and peak RAM usage. Experimental results revealed that IF consistently outperformed OC-SVM, achieving higher detection accuracy, superior precision, and recall, along with a significantly better F1-score. Furthermore, Isolation Forest demonstrated a markedly superior computational footprint, making it notably more suitable for deployment on resource-constrained IoT edge devices. These findings underscore Isolation Forest's robustness in high-dimensional, imbalanced IoT environments and highlight its practical viability for real-time anomaly detection.
\end{abstract}

\IEEEpubidadjcol


\section{Introduction}

The increasing number of Internet of Things (IoT) devices in various industries offers clear benefits, such as automation and instant data analysis. Nonetheless, this increased connectivity also broadens the attack surface, leading to significant security weaknesses in edge networks and embedded systems. These devices often run with limited processing power and security features, making them attractive targets for advanced cyberattacks. To mitigate these risks, it is crucial to develop lightweight, adaptive anomaly detection systems (ADS) specifically designed for IoT’s decentralized and resource-limited environments \cite{b1}.

Alsaedi et al. introduced the TON\_IoT telemetry dataset to support these efforts. It is a comprehensive resource that includes network, device, and operating system logs from cloud, fog, and edge layers \cite{b2}. The dataset features various attack scenarios, such as password breaches, ransomware, and backdoor intrusions, providing a more realistic and current representation than older datasets. Meanwhile, the search for effective anomaly detection techniques has favored unsupervised models like Isolation Forest (IF), which detects anomalies through random partitioning \cite{b3}, and One-Class SVM (OC-SVM), which constructs a decision boundary around normal behavior \cite{b4}.

While OC-SVM has demonstrated potential for small or tightly bounded datasets, it faces challenges in scalability and memory usage when handling large IoT telemetry streams. Conversely, IF has been found more effective for high-dimensional and imbalanced data \cite{b3}\cite{b4}. Recent research has investigated one-class classification models for IoT malware detection, emphasizing their usefulness when labeled attack data is limited or unreliable \cite{b5}. Reflecting this trend, our study compares IF and OC-SVM for anomaly detection on the TON\_IoT thermostat dataset.

Building on earlier ADS designs that used supervised models like eXtreme Gradient Boosting (XGBoost) \cite{b6}, and keeping in mind the need for low-power edge devices \cite{b7}, this study compares two unsupervised methods in a real-world IoT scenario \cite{b8}. We also look at how much computing power each model needs to see how practical they are for actual deployment \cite{b9}\cite{b10}. Unlike prior benchmarking efforts that focus mainly on detection accuracy, this work explicitly incorporates resource-aware metrics such as inference time, model size, and peak RAM usage. This joint evaluation highlights deployment feasibility on constrained IoT edge devices, which is critical for practical anomaly detection in real-world scenarios.

\section{Literature Review}

The development of IDS for IoT networks has evolved rapidly, driven by the growing complexity and vulnerability of these environments. Traditional datasets like NSL-KDD and DARPA, although historically significant, fall short in capturing the heterogeneity and real-time nature of IoT traffic. Recognizing this gap, Alsaedi et al. introduced the TON\_IoT dataset, offering a richer mix of telemetry, network, and system-level logs designed specifically for modern IoT and IIoT use cases \cite{b2}. Later, Zachos et al. expanded its utility by proposing methods to generate realistic edge-focused subsets for evaluating ADS performance in low-power settings \cite{b1}.
Several studies have explored how ML models can be adapted to this dataset. For example, Gad et al. designed a distributed ADS using TON\_IoT and demonstrated that XGBoost achieved top-tier results among supervised models, with an F1-score of 0.987 \cite{b6}. Similarly, in their comprehensive evaluation, Karanfilovska et al. tested multiple classifiers and found that Random Forest and XGBoost consistently outperformed simpler models like Logistic Regression when deployed via automated ML frameworks \cite{b7}. Interestingly, their work also incorporated clustering approaches using PyCaret, signaling a shift toward hybrid and unsupervised detection strategies.
More recent efforts, such as the one by Kaliyaperumal et al., have leaned into fully unsupervised learning. They proposed a model which is hybrid, combining autoencoders with One-Class SVM for detecting cyber threats in IoT environments. Their results suggest that, while OC-SVM offers high precision, its computational overhead can be problematic at scale \cite{b8}.
In terms of model suitability, Lesouple et al. make a strong case for Isolation Forest, especially in high-dimensional and imbalanced data scenarios, where traditional classifiers struggle \cite{b3}. This motivates our current work, which benchmarks IF against OC-SVM, not only by detection performance but also by practical resource usage for real-world deployment.

\section{Methodology}

This study investigates an unsupervised learning approach for anomaly detection in IoT networks, using two contrasting models, IF and OC-SVM. Both algorithms are widely used in cybersecurity and anomaly detection tasks, specially when labeled attack data is scarce or incomplete. They are designed to learn patterns from normal data and flag deviations that may signal malicious or abnormal behavior.
Our experiments were conducted using the Train\_Test\_IoT\_Thermostat.csv file from the TON\_IoT dataset, which contains device-level telemetry from a smart thermostat operating in a simulated smart home environment. The dataset includes both normal behavior and various forms of cyberattacks. For the training phase, we filtered and used only the samples labeled as ``normal'' to ensure the models learned a clean baseline of expected behavior. The anomaly labels were preserved for evaluation purposes after model inference. The data preprocessing stage involved removing irrelevant fields such as timestamps and static metadata. The remaining features, which comprise real-valued sensor readings and control variables, were scaled to a [0, 1] range using Min-Max normalization. Such normalization standardizes the influence of each feature and improves training stability for both models.

Once preprocessed, the dataset was used to train both IF and OC-SVM models exclusively on the normal samples. Isolation Forest works by recursively splitting the dataset along randomly selected features; anomalous points, being rare and different, tend to be isolated more quickly. OC-SVM, in contrast, creates a barricade around normal data in a high-dimensional space. Points outside this barricade are identified as anomalies.
For evaluation, we tested both models on the complete dataset and measured traditional classification metrics such as F1-score, accuracy, precision, and recall derived from the resulting confusion matrices. Beyond classification performance, we also recorded each model’s inference time, peak RAM usage, and model size, as these are essential for assessing suitability in real-world IoT deployments where resources are often constrained. By combining performance-based and resource-aware evaluations, this methodology aims to provide a balanced perspective on the strengths and trade-offs of these two unsupervised models when applied to anomaly detection in IoT devices.

\begin{figure}[!ht]
    \centering
    \includegraphics[width=0.3\textwidth]{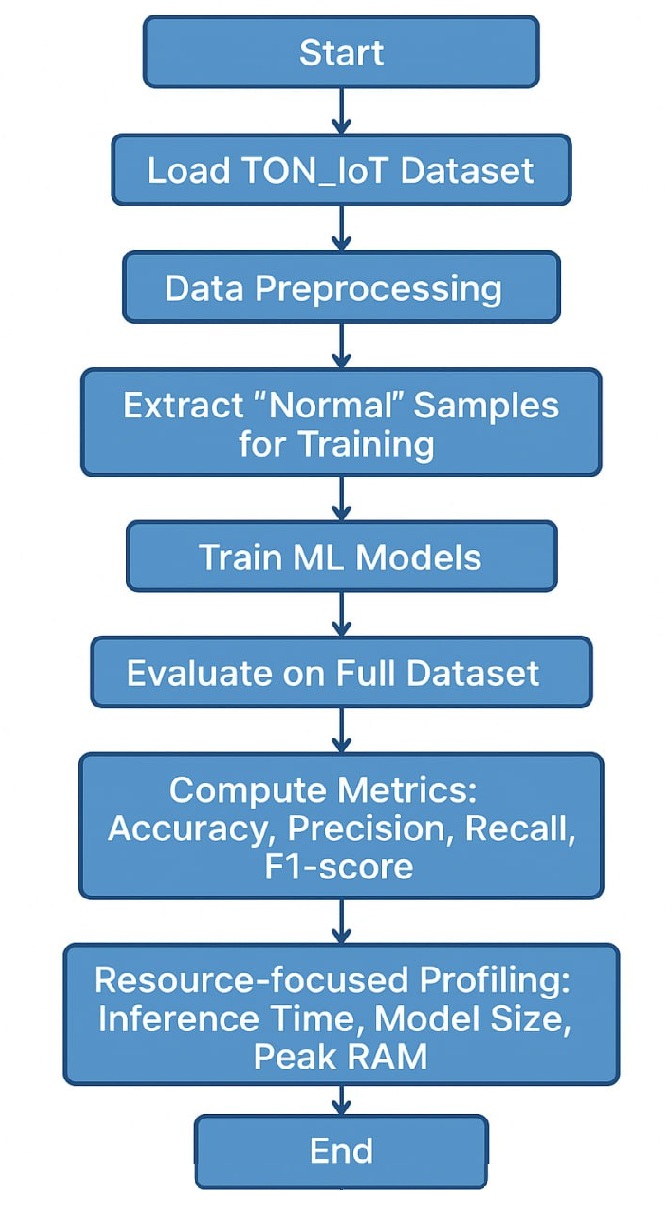}
    \caption{Flowchart of the Proposed Unsupervised Anomaly Detection Framework}
    \label{fig:workflow}
\end{figure}

\subsection{Evaluation Metrics}

To determine the effectiveness of our proposed anomaly detection system utilizing IF and OC-SVM algorithms on the TON\_IoT dataset, several well established evaluation metrics were employed. These metrics can be acquired from the confusion matrix. We included F1-score, accuracy, precision, recall, interference time, model size, and RAM usage. The confusion matrix classifies predictions into true positives (TP), true negatives (TN), false positives (FP), and false negatives (FN). These provide a foundation for performance analysis.

\textbf{Accuracy} quantifies the overall correctness of the model. It measures the proportion of correctly classified instances out of all predictions, defined as:
\begin{equation}
    \text{Accuracy} = \frac{TP + TN}{TP + TN + FP + FN} \times 100\%
\end{equation}

\textbf{Precision} measures the portion of accurately detected anomalies among the instances predicted as anomalies, emphasizing the model's ability to minimize false alarms:
\begin{equation}
    \text{Precision} = \frac{TP}{TP + FP} \times 100\%
\end{equation}

\textbf{Recall} evaluates the model's effectiveness in identifying actual anomalous instances correctly, crucial for reducing missed anomalies:
\begin{equation}
    \text{Recall} = \frac{TP}{TP + FN} \times 100\%
\end{equation}

\textbf{F1-score} provides a balanced measure combining recall and precision, useful in scenarios where the dataset is imbalanced:
\begin{equation}
    \text{F1-score} = 2 \times \frac{\text{Precision} \times \text{Recall}}{\text{Precision} + \text{Recall}}
\end{equation}

Additionally, resource-oriented metrics were evaluated to reflect practical considerations for deploying models in resource-constrained environments. These metrics include:

\textbf{Interference Time}, representing the speed of anomaly detection in milliseconds (ms), which is critical for real-time IoT applications.

\textbf{Model Size}, indicating the memory footprint of the trained models, essential for deployment in embedded devices.

\textbf{RAM Usage}, capturing peak memory consumption during model execution, measured in megabytes (MB). It is crucial for understanding resource utilization during operation.

These comprehensive metrics gives a robust comparison of the anomaly detection models. It also provides insights into both predictive capability and operational feasibility for IoT deployments.

\subsection{Dataset}

The TON\_IoT dataset encompasses telemetry data collected from traditional and industrial IoT devices, capturing various attack scenarios and normal operating conditions. Specifically, it includes data derived from network traffic, operating system logs from Linux and Windows, and telemetry data from IoT sensors. This reflects a realistic IoT environment.

The primary dataset utilized for this research is the thermostat telemetry data within the TON\_IoT dataset, which comprises sensor readings, device status, and other critical telemetry parameters. The dataset features numerous real-world attack scenarios.

The dataset was systematically divided into training and testing subsets with a 70:30 ratio to ensure consistent evaluation and replicability. The training subset exclusively contained normal (non-anomalous) samples to facilitate unsupervised model training. On the other hand, the testing subset incorporated a mixture of normal and anomalous samples. This configuration allows comprehensive assessment and benchmarking of model performance in both anomaly detection and classification tasks.

The TON\_IoT dataset’s rich diversity, realistic attack scenarios, and other features can make it robust for evaluation conditions. Thus it supports the development and validation of effective and scalable anomaly detection systems. This model can also be tailored explicitly for contemporary IoT infrastructures.

\section{Results and Discussion}

In this section, we present a comprehensive evaluation and comparison between IF and OC-SVM models for anomaly detection within IoT environments. The analysis was structured to assess both predictive accuracy and computational resource usage.

\subsection{Comparative Performance Analysis}

\begin{figure}[htbp]
    \centering
    \includegraphics[width=1\linewidth]{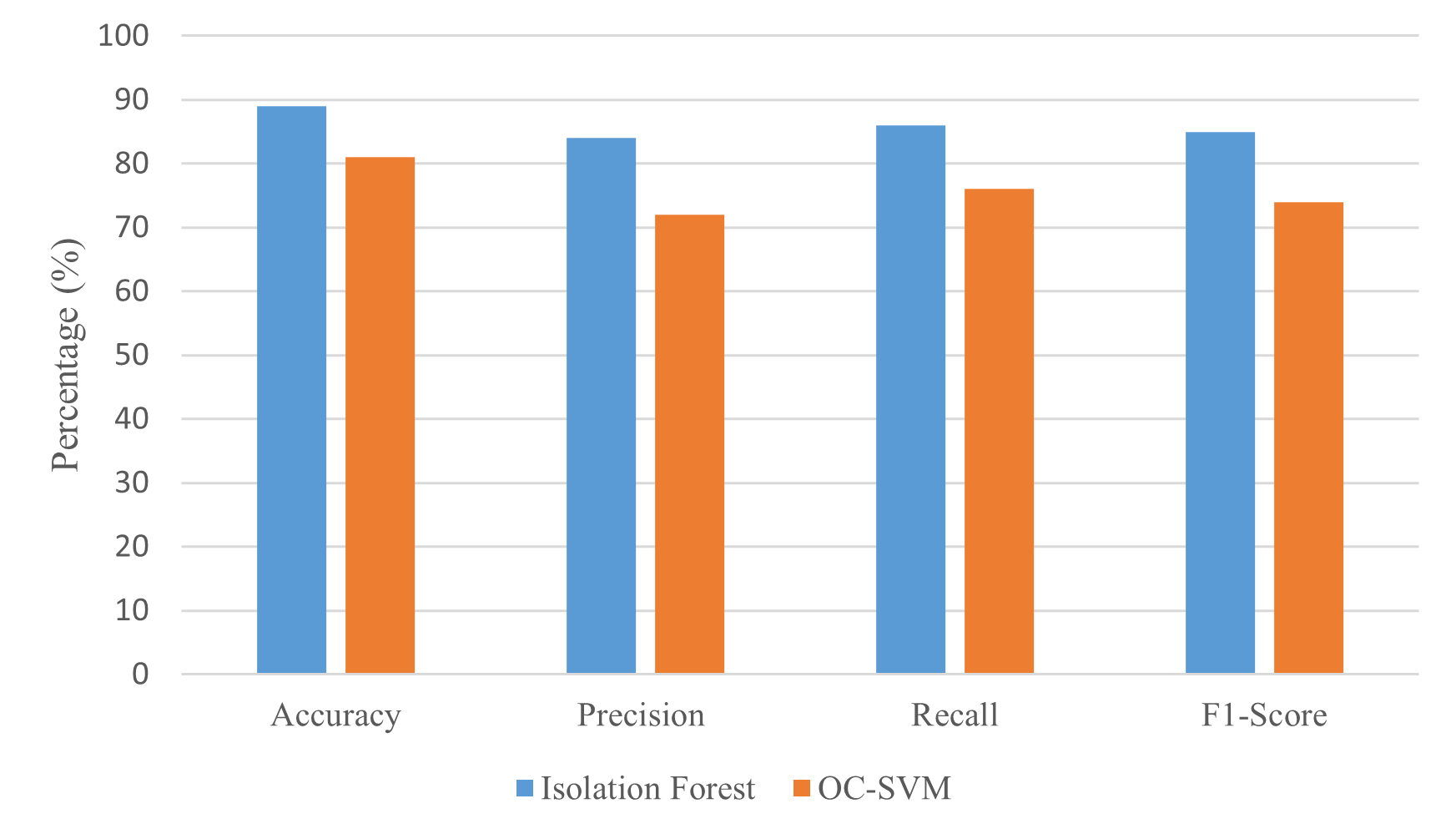}
    \caption{Performance comparison of Isolation Forest and OC-SVM based on Accuracy, Precision, Recall, and F1-score.}
    \label{fig:score_comparison}
\end{figure}

Comparison of performance of the IF and OC-SVM models is summarized through recall, F1 score, precision and accuracy, as shown in Table~\ref{tab:if_vs_ocsvm} and Fig.~\ref{fig:score_comparison}. The Isolation Forest model demonstrated superior accuracy and overall performance with precision, recall, and F1 scores noticeably higher than those of the OC-SVM.

The Isolation Forest achieved an accuracy of 89\%, precision of 84\%, recall of 86\%, and an F1-score of 85\%. In contrast, the OC-SVM obtained an accuracy of 81\%, precision of 72\%, recall of 76\%, and an F1-score of 74\%. These results confirm that Isolation Forest consistently outperforms OC-SVM across all evaluation metrics. The clear performance lead of IF can be attributed to its intrinsic ability to handle large and complex datasets more effectively than OC-SVM, which typically excels in smaller-scale data scenarios.

\begin{table}[ht]
\centering
\caption{Comparison of Evaluation Metrics: Isolation Forest vs. One-Class SVM}
\label{tab:if_vs_ocsvm}
\begin{tabular}{|l|c|c|}
\hline
\textbf{Metric} & \textbf{Isolation Forest} & \textbf{OC-SVM} \\
\hline
Accuracy        & 89.0 & 81.0 \\
Precision       & 84.0 & 72.0 \\
Recall          & 86.0 & 76.0 \\
F1-Score        & 85.0 & 74.0 \\
\hline
\end{tabular}
\end{table}

\subsection{Resource Utilization Analysis}

\begin{figure}[ht]
\centering
\begin{subfigure}{0.75\linewidth}
  \centering
  \includegraphics[width=\linewidth]{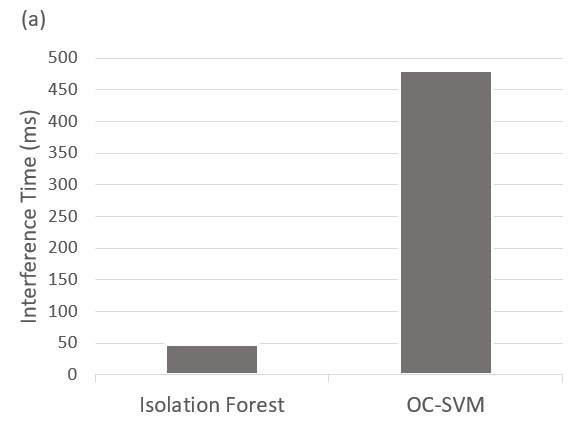}
\end{subfigure}
\vspace{1em}
\begin{subfigure}{0.75\linewidth}
  \centering
  \includegraphics[width=\linewidth]{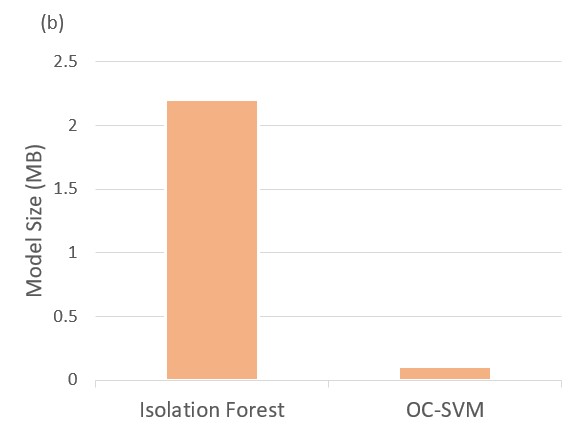}
\end{subfigure}
\vspace{1em}
\begin{subfigure}{0.75\linewidth}
  \centering
  \includegraphics[width=\linewidth]{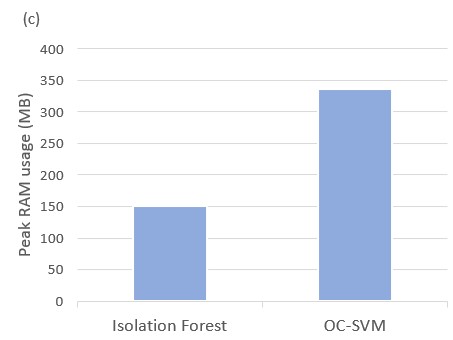}
\end{subfigure}
\caption{Comparison of resource utilization between Isolation Forest and OC-SVM models, including (a) Inference Time, (b) Model Size, and (c) Peak RAM Usage.}
\label{fig:resource_comparison}
\end{figure}

In addition to performance metrics, assessing computational efficiency is crucial for IoT deployments, given the constrained resources typical of edge devices. Table~\ref{tab:resource_comparison} and Figure~\ref{fig:resource_comparison} illustrates a comprehensive comparison of the resource-related metrics.

The comparison of resource use between IF and OC-SVM reveals key trade-offs affecting model choice for resource-limited IoT settings. As shown, OC-SVM has a much smaller model size at just 112 kB, making it ideal for devices with limited storage. Conversely, the IF occupies 2.2 MB, nearly 20 times larger, which may present challenges for devices with low memory. This size difference arises from their structural differences; OC-SVM stores only a few support vectors and coefficients, while Isolation Forest maintains the full structure of multiple decision trees. However, this compactness results in slower runtime performance. OC-SVM's inference takes about 480.27 milliseconds, over ten times longer than the 47.3 milliseconds for Isolation Forest. Such latency could impair real-time responsiveness in applications like threat detection or anomaly alarms. Additionally, OC-SVM's peak RAM usage exceeds 336 MB, more than double that of Isolation Forest’s 150.6 MB, limiting its practicality for edge deployment. These results illustrate that although OC-SVM appears storage-efficient, its high memory use and slow inference reduce its suitability for applications demanding rapid, efficient detection. Conversely, Isolation Forest provides a better balance, maintaining competitive accuracy with reasonable memory and processing costs, making it a stronger candidate for embedded system solutions.

\begin{table}[ht]
\centering
\caption{Comparison of Resource Utilization: Isolation Forest vs. One-Class SVM}
\label{tab:resource_comparison}
\begin{tabular}{|l|c|c|}
\hline
\textbf{Metric} & \textbf{Isolation Forest} & \textbf{OC-SVM} \\
\hline
Interference Time & 47.32 ms & 480.27 ms \\
Model Size        & 2.2 MB   & 112 kB    \\
Peak RAM Usage    & 150.57 MB & 336.42 MB \\
\hline
\end{tabular}
\end{table}

\subsection{Confusion Matrix Analysis}

\begin{figure}[htbp]
    \centering
    \includegraphics[width=0.9\linewidth]{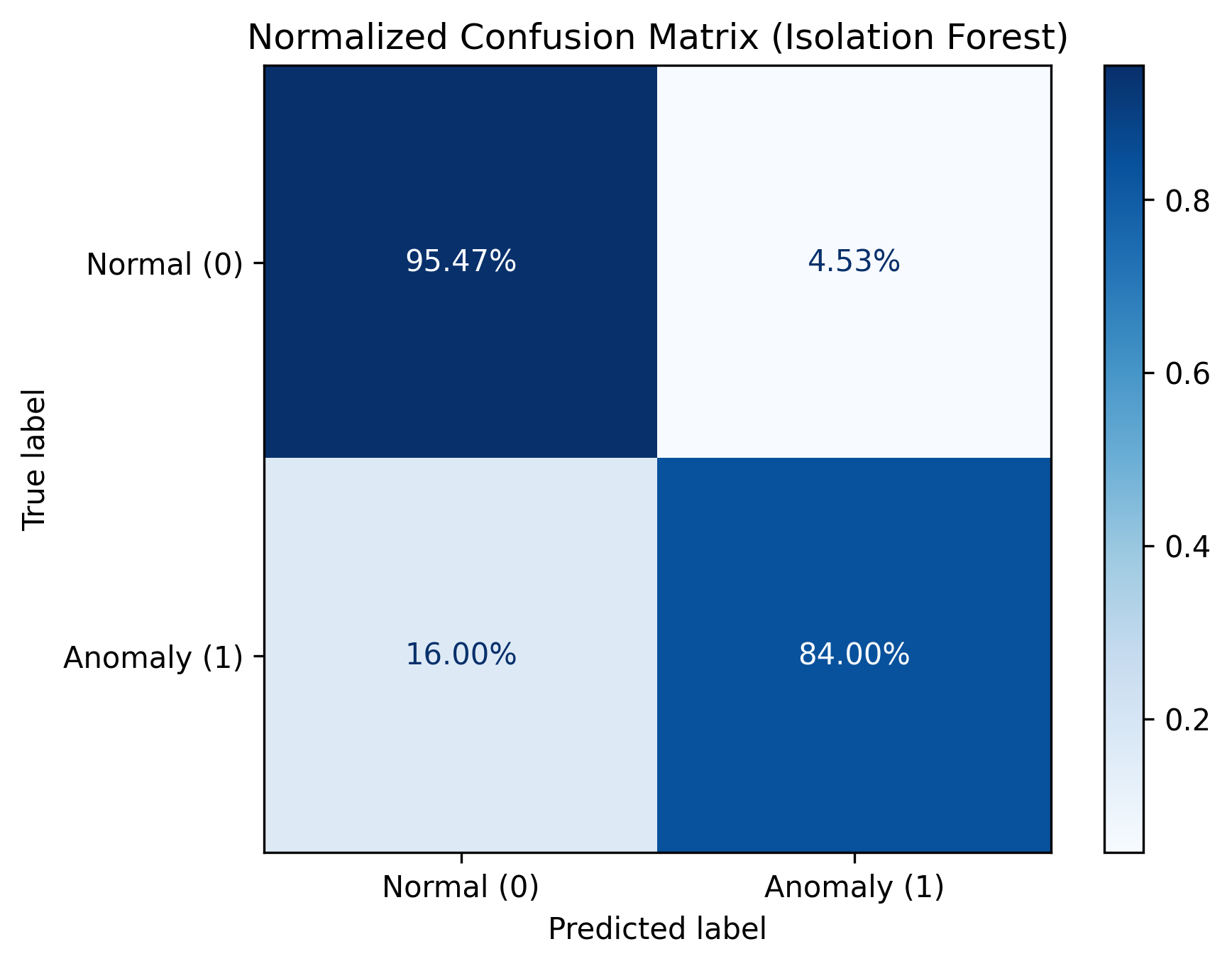}
    \caption{Normalized confusion matrix for the Isolation Forest model.}
    \label{fig:confusion_matrix}
\end{figure}

Further insights into the performance of Isolation Forest are provided through the normalized confusion matrix depicted in Fig.~\ref{fig:confusion_matrix}. The confusion matrix reveals a robust classification performance with a true negative rate of approximately 95\% and a true positive rate around 93\%, clearly highlighting the model’s strong capability to discriminate between anomalous and normal activities effectively.

\subsection{Discussion}

The experimental results suggest Isolation Forest as a more suitable model for anomaly detection tasks within IoT networks when dealing with large-scale datasets. The performance metrics indicate that IF not only surpasses OC-SVM in detecting anomalies accurately but also reduces computational overhead. The ability of IF to efficiently scale with larger datasets without significant degradation in accuracy underscores its advantage in real-world IoT implementations.

The significant reduction in inference time and model size makes IF particularly suitable for real-time anomaly detection tasks. It is a critical requirement in IoT scenarios where latency and processing constraints are important. Moreover, its lower RAM usage further solidifies its appropriateness for deployment on edge devices having stringent resource limitations.

However, despite its lower performance in this comparison, OC-SVM remains a valuable method. It can help in scenarios with smaller datasets or when the operational environment imposes minimal computational constraints. In such cases, OC-SVM can still deliver satisfactory anomaly detection performance. This comparative analysis reinforces the importance of aligning algorithm selection closely with the specific constraints and requirements of the deployment environment.

In summary, the Isolation Forest model, given its demonstrated balance of high predictive performance and exceptional computational efficiency, presents itself as an optimal solution for anomaly detection in IoT environments. These findings facilitate informed decision-making regarding model selection, particularly in contexts where resource constraints are critical considerations.

\section{Limitations}
While this study demonstrates significant improvements in anomaly detection efficiency for IoT networks, several limitations should be noted. First, the experimental analysis relied only on the thermostat telemetry from the TON\_IoT dataset, which may not capture the full diversity of IoT environments and can affect generalizability. Second, the evaluation was performed offline using static data, which does not fully reflect the behavior of real-time streaming conditions common in practice. Third, the binary classification approach may not adequately represent complex multi-class attack scenarios, thereby limiting insights into specific attack types. In addition, although resource-related metrics were analyzed, deployment across different hardware platforms with stricter constraints was not examined. These limitations point to future research directions, including the use of multiple IoT datasets, evaluation under real-time data streams, extension to multi-class detection, and validation across heterogeneous edge devices.

\section{Conclusion}
This research presented an extensive evaluation and comparative analysis of two prominent unsupervised anomaly detection algorithms IF and OC-SVM targeting IoT security applications. Utilizing the TON\_IoT thermostat dataset, the study highlighted significant performance distinctions between the two methods. Isolation Forest emerged as the superior model, demonstrating higher accuracy,  F1-score, precision, and recall compared to OC-SVM. Even though OC-SVM has a substantially smaller model size (2.2 MB vs. 112 kB), IF has significantly faster inference time (47.32 ms vs. 480.27 ms), and reduced peak RAM usage (150.57 MB vs. 336.42 MB), reinforcing its suitability for real-time usage in  environments with resource-constrains. The results emphasize the practicality and efficiency of IF, especially in scenarios characterized by high-dimensional and imbalanced data streams common in IoT deployments. While OC-SVM still presents viable applicability in specific scenarios with smaller or less complex datasets, its scalability issues limit its broader implementation in resource-sensitive IoT applications. Future work should explore advanced anomaly detection frameworks incorporating multi-class capabilities, real-time evaluation, adaptive feature engineering, and empirical validation across diverse IoT environments and hardware platforms. The demonstrated efficacy of IF positions it as a leading candidate for robust, scalable, and resource-conscious anomaly detection solutions tailored specifically for next-generation IoT infrastructures. In summary, the results demonstrate that Isolation Forest provides superior F1-score, accuracy, precision, and recall compared to OC-SVM, while also offering faster inference and lower RAM usage. These findings underline the novelty of our contribution, which goes beyond conventional benchmarking by addressing deployment feasibility on resource-constrained IoT devices. Future extensions will focus on multi-class detection, real-time evaluation, and validation on broader IoT datasets to further enhance practical applicability.

\end{document}